\documentclass[9pt,twocolumn]{extarticle}
\usepackage{modernstyle}

\usepackage[round]{natbib}

\usepackage{times}
\usepackage{soul}
\usepackage{url}
\usepackage[hidelinks]{hyperref}
\usepackage[utf8]{inputenc}
\usepackage{caption}
\usepackage{graphicx}
\usepackage{amsmath}
\usepackage{amsthm}
\usepackage{booktabs}
\usepackage{algorithm}
\usepackage{dirtree}

\usepackage{multirow}
\usepackage{makecell}

\usepackage{algpseudocode}
\algnewcommand{\LineComment}[1]{\State \(\triangleright\) \textit{#1}} %
\algrenewcommand{\textproc}[1]{#1} %
\algnewcommand\Input[1]{\State \textbf{Input:} #1}%
\algnewcommand\Output[1]{\State \textbf{Output:} #1}%

\usepackage{xcolor}
\definecolor{okgreen}{RGB}{2, 137, 59}
\definecolor{errorred}{RGB}{165, 25, 0}
\definecolor{strs}{HTML}{1F4E79}
\definecolor{kwrds}{HTML}{7A1E1E}
\definecolor{numb}{HTML}{3F6B3F}

\usepackage[T1]{fontenc}
\usepackage{listings}
\lstdefinelanguage{json}{
    basicstyle=\normalfont\ttfamily,
    commentstyle=\color{strs},
    stringstyle=\color{kwrds},
    numbers=left,
    numberstyle=\scriptsize,
    stepnumber=1,
    numbersep=8pt,
    showstringspaces=false,
    breaklines=true,
    frame=lines,
    string=[s]{"}{"},
    comment=[l]{:\ "},
    morecomment=[l]{:"},
    literate=
        *{0}{{{\color{numb}0}}}{1}
         {1}{{{\color{numb}1}}}{1}
         {2}{{{\color{numb}2}}}{1}
         {3}{{{\color{numb}3}}}{1}
         {4}{{{\color{numb}4}}}{1}
         {5}{{{\color{numb}5}}}{1}
         {6}{{{\color{numb}6}}}{1}
         {7}{{{\color{numb}7}}}{1}
         {8}{{{\color{numb}8}}}{1}
         {9}{{{\color{numb}9}}}{1}
         {true}{{{\color{numb}\ \ true}}}{1}
}

\usepackage{dsfont}
\usepackage{pifont}
\usepackage{marvosym}
\newcommand{\cmark}{\textcolor{okgreen}{\ding{51}}}%
\newcommand{\xmark}{\textcolor{errorred}{\ding{55}}}%

\newcommand{\Args}{\mathcal{A}}
\newcommand{\Atts}{\mathcal{R}^-}

\newcommand{\Supps}{\mathcal{R}^+}
\newcommand{\BS}{\ensuremath{\tau}}
\newcommand{\SF}{\ensuremath{\sigma}}

\newcommand{\QBAF}{\mathcal{Q}}

\usepackage{fancyhdr}
\fancypagestyle{firstpage}{%
	\fancyhf{}
	\setlength{\headsep}{0.3in}
	\cfoot{\thepage}
}
\fancypagestyle{otherpages}{%
	\fancyhf{}
	\setlength{\headsep}{0.3in}
	\rhead{Argumentation for Explainable and Globally Contestable Decision Support with LLMs}
	\lhead{A. Dejl, M. Williams and F. Toni}
	\cfoot{\thepage}
}

\title{Argumentation for Explainable and Globally Contestable\\ Decision Support with LLMs}

\author {
	Adam Dejl$^1$, Matthew Williams$^2$ and Francesca Toni$^1$\\
	\large
	$^1$Department of Computing, Imperial College London \\
	$^2$Department of Surgery \& Cancer, Imperial College London \\
	\medskip
	\{adam.dejl18, matthew.williams, ft\}@imperial.ac.uk
}

\begin{document}

\pagestyle{otherpages}
\maketitle

\begin{abstract}
    Large language models (LLMs) exhibit strong general capabilities, but their deployment in high-stakes domains is hindered by their opacity and unpredictability. Recent work has taken meaningful steps towards addressing these issues by augmenting LLMs with post-hoc reasoning based on computational argumentation, providing faithful explanations and enabling users to contest incorrect decisions. However, this paradigm is limited to pre-defined binary choices and only supports local contestation for specific instances, leaving the underlying decision logic unchanged and prone to repeated mistakes. In this paper, we introduce ArgEval, a framework that shifts from instance-specific reasoning to structured evaluation of general decision options. Rather than mining arguments solely for individual cases, ArgEval systematically maps task-specific decision spaces, builds corresponding option ontologies, and constructs general argumentation frameworks (AFs) for each option. These frameworks can then be instantiated to provide explainable recommendations for specific cases while still supporting global contestability through modification of the shared AFs. We investigate the effectiveness of ArgEval on treatment recommendation for glioblastoma, an aggressive brain tumour, and show that it can produce explainable guidance aligned with clinical practice\footnote{Our code is available at \url{https://github.com/adamdejl/argeval}.}.
\end{abstract}

\begin{figure}[htb]
    \centering
    \includegraphics[width=0.93\linewidth]{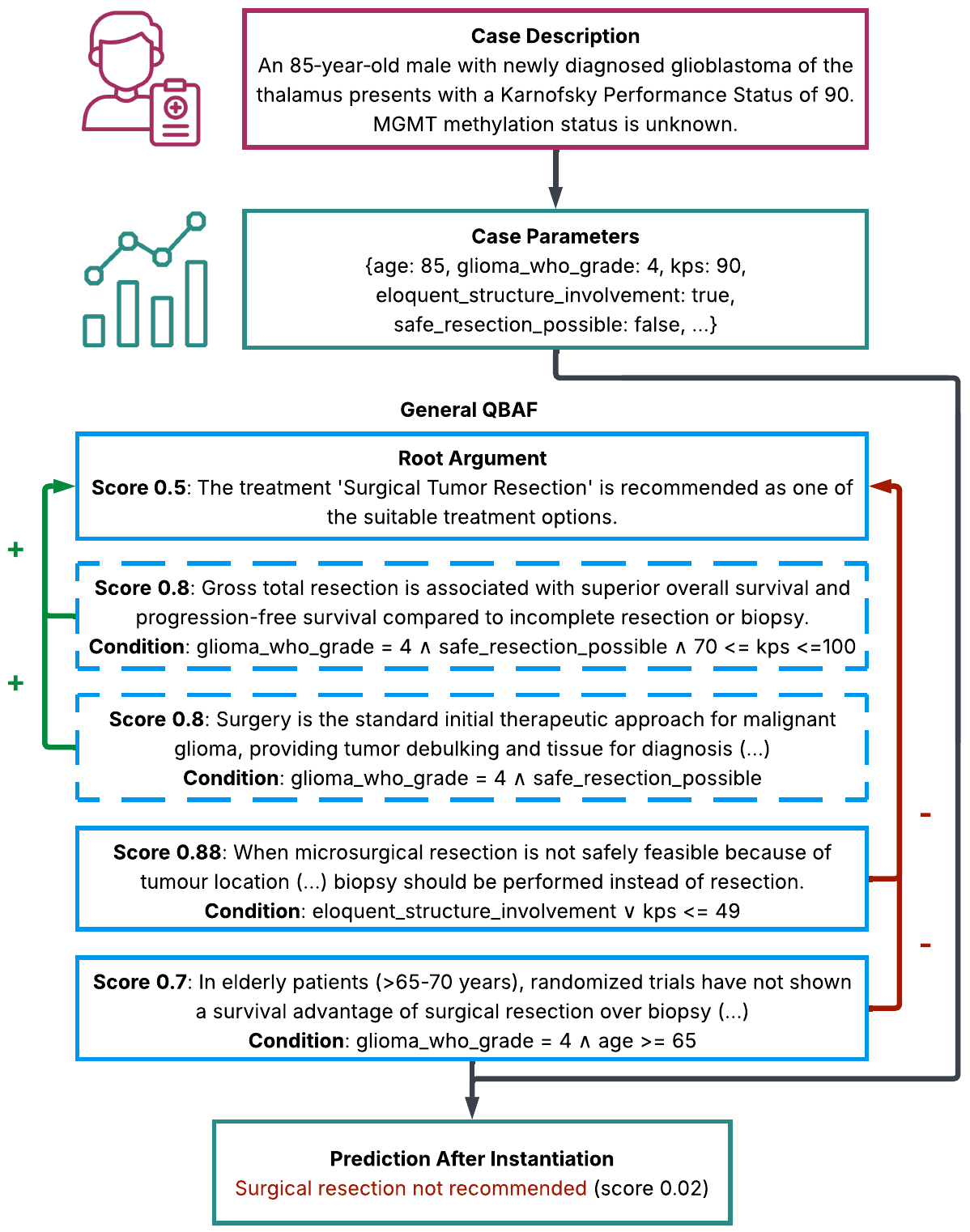}
    \caption{Illustration of ArgEval inference for one of the treatment options for glioblastoma\protect\footnotemark. {\bf Case Parameters} extracted from the {\bf Case Description} are used to instantiate the {\bf General QBAF} associated with the given option ({\bf Root Argument}), removing the dashed nodes whose conditions are not satisfied. The {\bf Prediction After Instantiation} is obtained from 
    the instantiated framework, which also serves as a faithful explanation.}
    \label{fig:argeval-inference}
\end{figure}

\footnotetext{Icons from Noun Project (\href{https://thenounproject.com/icon/patient-6651912/}{Patient} by Alzam and \href{https://thenounproject.com/icon/document-5079157/}{Data} by Anna Riana, \href{https://creativecommons.org/licenses/by/3.0/deed.en}{CC BY 3.0}).}

\begin{figure*}[tb]
    \centering
    \includegraphics[width=0.84\textwidth]{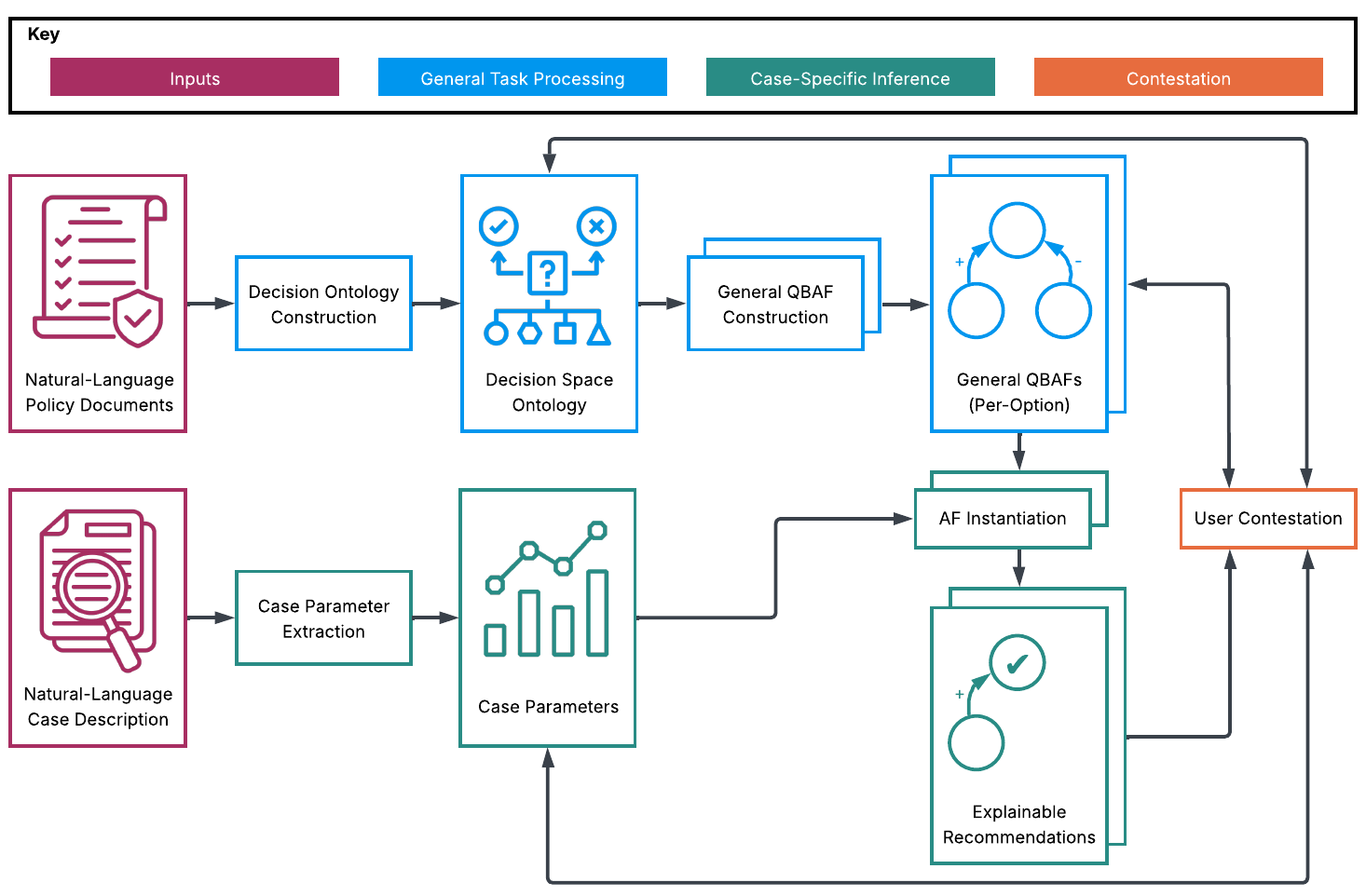}
    \caption{Overview of the ArgEval pipeline\protect\footnotemark. Top: given natural-language policy documents that specify general criteria for decision-making in a certain domain, ArgEval builds a decision-space ontology and constructs general QBAFs for each candidate decision in the ontology. Bottom: at inference time, the general QBAF is instantiated with the parameters of a specific case, providing faithfully explainable recommendations. Users can contest the decision-space ontology, the general QBAFs, the extracted case parameters and the general parameter schema specifying the properties to be extracted, in response to incorrect recommendations or explanations produced by the model.}
    \label{fig:argeval-pipeline}
\end{figure*}

\footnotetext{Icons from Noun Project (\href{https://thenounproject.com/icon/policy-8208982/}{Policy} by Puspito, \href{https://thenounproject.com/icon/decision-5223502/}{Decision} by Kamin Ginkaew, \href{https://thenounproject.com/icon/report-analysis-6505267/}{report analysis} by kartini 1, \href{https://thenounproject.com/icon/patient-6651912/}{Patient} by Alzam and \href{https://thenounproject.com/icon/data-8259289/}{Data} by Anna Riana, \href{https://creativecommons.org/licenses/by/3.0/deed.en}{CC BY 3.0}).}

\section{Introduction}

Trained on large corpora of general textual data as well as specific problem-solving tasks, large language models (LLMs) have demonstrated strong performance in a variety of settings \citep{bubeck-2023-sparks-gai,vanveen-2024-llms-outperform-summaries,luo-2025-llms-surpass-neuroscience,mcduff-2025-differential-diagnosis,bermejo-2025-llms-outperform-human-coders}. However, due to their reliance on stochastic next-token prediction, LLMs can also be notoriously unreliable, as evidenced by issues such as hallucinations \citep{huang-survey-hallucinations} and omissions of critical information \citep{busch-2025-llm-applications-patient-care}. These issues pose a substantial obstacle to safely using LLMs in high-stakes domains. Importantly, the inherent opacity of LLMs makes it impossible to faithfully explain their outputs or to reliably redress issues. While methods such as chain-of-thought \citep{wei-2022-cot} or verbalised explanations can provide some degree of rationalization, they have been found to be unfaithful, i.e., inaccurate in describing the true internal reasoning of models \citep{chen-2025-reasoning-unfaithful}. 

In this work, we introduce \emph{ArgEval}, a new framework for decision support with LLMs that leverages argumentative reasoning to produce explainable predictions that can be globally contested. Like existing approaches combining LLMs with argumentation~\citep{freedman-2025-argllms,kevin,arg-rag,deniz-2025-multi-agent}, ArgEval uses an LLM for mining arguments for and against a particular decision as well as estimating numerical scores indicating their intrinsic strength (also called \emph{base scores}). The arguments and scores are then used to form a \emph{quantitative bipolar argumentation framework} (QBAF), enabling deterministic inference via gradual semantics \citep{baroni-2019-qbafs}. Applying the semantics yields a set of final argument strengths, which indicate the degree of acceptability of each argument after considering its interactions with the other arguments in the QBAF. The strength of the argument representing the considered decision can then be used to make predictions, with the QBAF acting as an inherently faithful explanation for these predictions.

While sharing the use of QBAFs with prior works, ArgEval differs in several ways. Unlike ArgLLMs~\citep{freedman-2025-argllms,kevin} and like ArgRAG~\citep{arg-rag}, ArgEval uses external sources to guide the mining of QBAFs. However, these sources are used to compile a \emph{decision-space ontology} and  \emph{general QBAFs}, respectively representing the available decision options and general knowledge concerning the decision-making problem of interest, differing from ArgRAG, which only uses them to make predictions relating to specific instances. Unlike both ArgLLMs and ArgRAG, ArgEval is not restricted to binary questions (claims) but is applicable to open-ended decisions, in the spirit of Evaluative AI~\citep{miller-2023-evaluative-ai}.

Uniquely, when making predictions for specific cases, ArgEval instantiates the general QBAFs associated with each decision option rather than constructing these QBAFs from scratch (see illustration in Figure~\ref{fig:argeval-inference}). As with ArgLLMs and ArgRAG, the resulting instantiated QBAFs can be used as faithful explanations. Users can inspect these QBAFs and \emph{contest} them to correct any mistakes, either by changing argument base scores or by adding additional arguments. However, ArgEval shifts contestation from case-specific, \emph{local contestability}, as supported by ArgLLMs and ArgRAG, to \emph{global contestability}, potentially affecting other future cases. Since the arguments in the instantiated QBAF directly correspond to arguments in the general QBAF, any changes to those arguments will directly affect predictions for any cases satisfying their applicability conditions. Apart from the QBAFs, users may also inspect and adjust the decision-space ontology, the extracted case parameters used for QBAF instantiation and the general parameter schema specifying the properties to be extracted for each case.

To evaluate the effectiveness of ArgEval, we apply it to the task of recommending suitable treatments for different cases of glioblastoma based on relevant clinical guidelines. We show that variants of ArgEval perform competitively against the reasoning LLM baseline and ArgLLMs-O, an adapted version of ArgLLMs using the extracted decision-space ontology for supporting open-ended decisions. Importantly, ArgEval achieves this performance using a fraction of the computational cost associated with the other methods while providing a significant qualitative advantage in the form of global contestability.

To summarise, our key contributions are as follows:

\begin{itemize}
    \item We propose ArgEval (Figure~\ref{fig:argeval-pipeline}), a new framework for decision support that provides faithful explainability and global contestability through argumentation.
    \item We apply ArgEval to treatment recommendation for glioblastoma brain tumours, showing that it exhibits competitive performance compared to other methods at a fraction of the inference cost.
    \item We demonstrate the benefits of global contestability in a case study where contestation on a single sample substantially improves the performance of the model.
\end{itemize}

\begin{table}[!tb]
     \caption{Qualitative comparison between ArgEval and other methods. (\cmark) in the third row signifies that local contestability for LLMs relies on their opaque mechanisms rather than being fully transparent. A higher number of \Lightning\ symbols denotes faster inference. More details regarding the listed properties are provided in Section \ref{sec:argeval-properties}.}
     \label{tab:argeval-comparison}
     \centering
     \footnotesize
     \begin{center}
     \begin{tabular}{ c c c c c } 
        \toprule
        \multirow{2}{*}{\textbf{Property}} & \multicolumn{4}{c}{\textbf{Method}} \\
        \cmidrule(r){2-5}
        & \textbf{LLMs} & \makecell{\textbf{ArgLLMs} \\ \textbf{ArgRAG}} & \textbf{ArgLLMs-O} & \textbf{ArgEval} \\
        \midrule
        Open-Ended Decisions & \cmark & \xmark & \cmark & \cmark \\
        Faithful Explainability & \xmark & \cmark & \cmark & \cmark \\
        Local Contestability & (\cmark) & \cmark & \cmark & \cmark \\
        Global Contestability & \xmark & \xmark & \xmark & \cmark \\
        Inference Speed & \Lightning \Lightning & \Lightning & \Lightning & \Lightning \Lightning \Lightning \\
        \bottomrule
     \end{tabular}
     \end{center}
\end{table}

\section{Related Work}

\paragraph{Argumentation and LLMs}
The use of argumentation in combination with LLMs is advocated by several. Argumentative LLMs (ArgLLMs) \citep{freedman-2025-argllms} have been recently proposed as a step towards making LLMs more reliable and responsive to user-driven error corrections (i.e., contestations), in the form of a paradigm combining general LLM knowledge and capabilities with formal argumentative reasoning. We adopt the same philosophy as ArgLLMs and use them as a baseline to compare against, but we are not restricted to binary decisions (such as determining whether a certain claim is truthful or not) or local contestation.

In a similar spirit, ArgRAG~\citep{arg-rag} uses retrieval-augmented generation (RAG) with external sources for generating QBAFs, making binary judgements about claim validity. Like ArgRAG, we inform our QBAF generation with external sources, but these are compiled into a structured ontology instead of being retrieved ad hoc as in ArgRAG. Moreover, similarly to ArgLLMs and unlike ArgRAG, we restrict the QBAF extraction to acyclic graphs.

Both ArgLLMs and ArgRAG were originally proposed for claim verification, and have since been adapted to other use-cases, such as judgemental forecasting \citep{deniz-2025-multi-agent}. In addition to approaches that extract full QBAFs, some works use LLMs for mining attack and support relations between pieces of text treated as arguments, as in \cite{deniz-2025-multi-agent}, or arguments, their components and/or relations between them, as in~\cite{cabessa-2025-argument-mining}.

\paragraph{Argumentation Schemes and  Ontologies}
Our general QBAFs, such as those used in the glioblastoma treatment recommendation experiment, can be informed by (existing or new) argumentation schemes with critical questions \citep{walton-argument-schemes}, to be instantiated for specific cases of interest. 
 This approach is inspired by \cite{hamed-2025-bias}, though their method does not use LLMs or any ontology to instantiate the argument schemes. Our use of ontologies for QBAF instantiation is related to \cite{antonio-2025-reviews}, which, however, uses them directly in combination with text mining rather than to form general QBAFs.

\paragraph{Contestable AI}
Contestability is increasingly advocated as important in  human-in-the-loop AI systems \citep{lyons2021conceptualising}, with argumentation indicated as particularly suitable to support AI contestability \citep{leofante-2024-contestable-ai,dignum-2025-contestable-ai}. However, existing approaches to contestability via argumentation, notably \cite{freedman-2025-argllms,xiang-2025-contestability-arg}, focus on local contestability, for single input-output pairs. \cite{contestABAL} propose a  form of global contestability, where learnt argumentation-based models are contested, but these are structured argumentation frameworks rather than QBAFs.  

\paragraph{Faithfulness of Explanations}
While often proposed as a simple way of explaining LLM reasoning, chain-of-thought (CoT) has been found to be unfaithful to the true reasoning process of models, failing to mention factors that affect the outputs \citep{chen-2025-reasoning-unfaithful}. Thus, CoT is insufficient for faithful interpretability \citep{barez-2025-cot-not-explainability}. Some works, e.g. \cite{oana-faithfulness2023}, study the faithfulness of various post-hoc explanation methods for LLMs' outputs, e.g. saliency maps and counterfactuals. Other works, e.g. \cite{oana-faithfulness}, propose faithfulness metrics for explanations generated as reasoning traces by LLMs. ArgEval, like ArgLLMs and ArgRAG, instead opts for drawing explanations from generated QBAFs, in the spirit of \cite{xai-survey},
which are faithful by construction.

\section{Preliminaries}
\label{sec:preliminaries}
A \emph{QBAF}
is a quadruple $\mathcal{Q}=\left\langle\mathcal{A}, \mathcal{R}^{-}, \mathcal{R}^{+}, \tau \right\rangle$ where:
\begin{itemize}
    \item $\mathcal{A}$ is a finite set of \emph{arguments};
    \item $\mathcal{R}^{-} \subseteq \mathcal{A} \times \mathcal{A}$ is a binary \emph{attack} relation;
    \item $\mathcal{R}^{+} \subseteq \mathcal{A} \times \mathcal{A}$ is a binary \emph{support} relation;
    \item $\mathcal{R}^{-} \cap \mathcal{R}^{+} = \emptyset$;
    \item $\tau: \mathcal{A} \rightarrow  [0,1]$ is a \emph{base score function}.
\end{itemize}

Arguments in QBAFs may be evaluated by \emph{gradual semantics}~\citep{baroni-2019-qbafs},  a total function $\sigma: \mathcal{A} \rightarrow [0,1]$ which, for any $\alpha \in \mathcal{A}$, assigns a \emph{strength} $\sigma(\alpha)$ to $\alpha$. While $\BS(\alpha)$ can be seen as the \emph{intrinsic strength} for $\alpha$, the strength $\SF(\alpha)$ can be seen as ``dialectical'', following the debate captured by $\Atts$ and $\Supps$. For a given QBAF $\mathcal{Q}$, we let $\sigma_{\QBAF}(\alpha)$ denote the strength of $\alpha \in \mathcal{A}$ in $\mathcal{Q}$.

In this paper we adopt the following restrictions on QBAFs.
Each QBAF is a tree, with the root being an argument representing a decision option (similarly to ArgLLMs, where the root is a claim).
Each such tree has either \emph{depth 1} or \emph{depth 2}. At depth 1, the root is attacked and/or supported by a finite number of arguments; at depth 2, each of those arguments can be attacked and/or supported by any number of further arguments. Unlike ArgLLMs, which allow at most one attacker and one supporter for each argument, our QBAFs can have arbitrary breadth. 

For the experiments, we use the \emph{discontinuity-free quantitative argumentation debate} (DF-QuAD) gradual semantics~\citep{RagoTAB16}, which is defined as follows. For any $\alpha \in \Args$ with $n\!\geq \! 0$  attackers with strengths $v_1, \ldots, v_n$, $m\geq 0$  supporters with strengths $v_1', \ldots, v_m'$ and $\BS(\alpha)=v_{0}$, $\SF(\alpha)=\mathcal{C}(v_{0}, \mathcal{F}(v_{1}, \ldots, v_{n}), \mathcal{F}(v_{1}', \ldots, v_{m}'))$, where
\begin{itemize} 
\item for $v_{a}=\mathcal{F}(v_{1}, \ldots, v_{n})$ and $v_{s}=\mathcal{F}(v_{1}', \ldots, v_{m}')$:
if $v_a=v_s$ then $\mathcal{C}(v_{0}, v_{a}, v_{s})=v_0$; else if $v_a > v_{s}$ then $\mathcal{C}(v_{0}, v_{a}, v_{s})=v_{0} - (v_{0}\cdot|v_{s}-v_{a}|)$; otherwise $\mathcal{C}(v_{0}, v_{a}, v_{s})=v_{0} + ((1-v_{0})\cdot|v_{s}-v_{a}|)$; and
\item 
given $n$ arguments with strengths $v_{1}, \ldots, v_{n}$, if $n=0$ then $\mathcal{F}(v_{1}, \ldots, v_{n})= 0$, otherwise $\mathcal{F}(v_{1}, \ldots, v_{n})= 1 - \prod_{i=1}^n (|1-v_{i}|)$.
\end{itemize}

\noindent {\em Argument schemes}
~\citep{walton-argument-schemes} are debate templates composed of major/minor premises, conclusions and critical questions on the minor premises' validity. The schemes guide how  arguments are built and the critical questions point to ways to attack them.

\section{ArgEval}
\label{sec:argeval}

The ArgEval pipeline (illustrated in Figure \ref{fig:argeval-pipeline}) consists of two main stages: general task processing, which is responsible for mapping the task domain and identifying general rules for making decisions in that domain, and case-specific inference, which provides decision recommendations for specific instances. We describe the specifics of each stage below.

\subsection{General Task Processing}

\subsubsection{Decision-Space Ontology Construction} The initial step of the general task processing automatically maps the decision space associated with the given domain, constructing a structured ontology of possible decision options. As source material for constructing the ontology, the pipeline relies on a corpus of natural-language policy documents $\mathcal{D}$ describing the general procedures for making decisions on the given task. For example, when aiming to provide decision support for choosing between different treatments for a certain medical condition, $\mathcal{D}$ may consist of the relevant clinical guidelines. Each document $d \in \mathcal{D}$ is decomposed into a set of text chunks $\mathcal{T}_d$, split semantically according to the document sections. Optionally, these chunks can be filtered according to their relevance to the considered task using an LLM, yielding a filtered set $\mathcal{T}'_d$ (or $\mathcal{T}'_d = \mathcal{T}_d$ when filtering is not applied).

\begin{algorithm}[tb]
\caption{Decision-Space Ontology Construction}
\label{alg:ontology-construction}
\textbf{Input}: A set of documents $\mathcal{D}$ and a set of text chunks $\mathcal{T}'_d$ for each document $d \in \mathcal{D}$ \\
\textbf{Output}: Constructed ontology $\mathcal{O}$
\begin{algorithmic}[1]
    \State Initialise $\mathcal{E} \gets \emptyset$, $\mathcal{T} \gets \emptyset$, $\mathcal{H} \gets \emptyset$, $\mathcal{S} \gets \emptyset$
    \For{$d \in \mathcal{D}$}
        \For{$t \in \mathcal{T}'_d$}
            \LineComment{Extract chunk entities and hierarchy relations}
            \State $\langle \mathcal{E}_t, \mathcal{H}_t \rangle \gets$ \texttt{mine\_ontology}($\mathcal{E}$, $\mathcal{H}$, $t$)
            \State $\mathcal{E} \gets \mathcal{E} \cup \mathcal{E}_t$
            \State $\mathcal{T} \gets \mathcal{T} \cup \{ t \}$
            \State $\mathcal{H} \gets \mathcal{H} \cup \mathcal{H}_t$
            \State $\mathcal{S} \gets \mathcal{S} \cup \{(e, t) \mid e \in \mathcal{E}_t\}$
        \EndFor
    \EndFor
    \State $\mathcal{O} \gets$  $\langle \mathcal{E}, \mathcal{T}, \mathcal{H}, \mathcal{S} \rangle$
    \State \Return $\mathcal{O}$
\end{algorithmic}
\end{algorithm}

\begin{algorithm}[tb]
\caption{General QBAF Construction}
\label{alg:qbaf-construction}
\textbf{Input}: 
    Ontology $\mathcal{O} = \langle \mathcal{E}, \mathcal{T}, \mathcal{H}, \mathcal{S} \rangle$, 
    mining depth $d$, 
    argument scheme $s_\text{arg}$ (optional), 
    boolean flag \texttt{score\_root} \\
\textbf{Output}: A set of general QBAFs $\mathbf{G} = \{\mathcal{G}_e\}_{e \in \mathcal{E}}$ and global parameter schema $\Pi$
\begin{algorithmic}[1]
    \State Initialise $\mathbf{G} \gets \emptyset$, $\Pi \gets \emptyset$
    
    \For{$e \in \mathcal{E}$}
        \State Initialise $\tau_e \gets \emptyset$, $\chi_e \gets \emptyset$
        \State $\mathcal{T}_e \gets \{t \in \mathcal{T} \mid (e, t) \in \mathcal{S}\}$
        
        \LineComment{1. Mine a bipolar argumentation framework}
        \LineComment{\phantom{1.} with a natural-language condition function $\chi_e^N$}
        \State $\langle \mathcal{A}_e, \mathcal{R}^-_e, \mathcal{R}^+_e, \chi_e^N \rangle \gets \texttt{mine\_baf}(e, \mathcal{T}_e, d, s_\text{arg})$
        
        \For{$a \in \mathcal{A}_e$}
            \LineComment{2. Estimate base scores}
            \If{$\texttt{is\_root}(a)$ \textbf{and not} \texttt{score\_root}}
                \State $\tau_e(a) \gets 0.5$
                \State $\chi_e(a) \gets \top$ \Comment{\textit{Use tautology for root}}
            \ElsIf{$\texttt{is\_root}(a)$}
                \State $\tau_e(a) \gets \texttt{score}(a, \mathcal{T}_e)$
                \State $\chi_e(a) \gets \top$ \Comment{\textit{Use tautology for root}}
            \Else
                \State $p \gets \texttt{get\_parent}(a, \mathcal{R}^-_e, \mathcal{R}^+_e)$
                \State \texttt{is\_sup} $\gets (a, p) \in \mathcal{R}^+_e$
                \State $\tau_e(a) \gets \texttt{score}(a, \mathcal{T}_e, p, \texttt{is\_sup}, \chi_e^N(a))$

                \LineComment{3. Formalise conditions, updating schema}
                \State $\langle \chi_e(a), \Pi \rangle \gets \texttt{form\_cond}(a, \chi_e^N(a), \Pi)$
            \EndIf
        \EndFor
        
        \State $\mathcal{G}_e \gets \langle \mathcal{A}_e, \mathcal{R}^-_e, \mathcal{R}^+_e, \tau_e, \chi_e \rangle$
        \State $\mathbf{G} \gets \mathbf{G} \cup \{\mathcal{G}_e\}$
    \EndFor
    
    \State \Return $\mathbf{G}, \Pi$
\end{algorithmic}
\end{algorithm}

The obtained chunks are then used to construct an ontology $\mathcal{O} = \langle \mathcal{E}, \mathcal{T}, \mathcal{H}, \mathcal{S} \rangle$ where $\mathcal{E}$ is a set of decision entities in the ontology, $\mathcal{T}$ is the associated set of source text chunks, $\mathcal{H} \subseteq \mathcal{E} \times \mathcal{E}$ is a hierarchy relation linking specific variants of decision options to their general categories and $\mathcal{S} \subseteq \mathcal{E} \times \mathcal{T}$ is a provenance relation linking each decision entity to all the relevant chunks in which it is mentioned.

\begin{figure}[!htb]
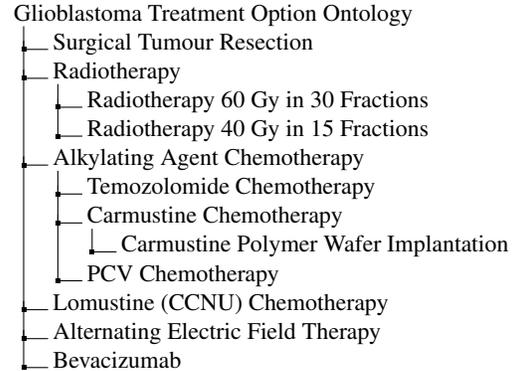

    \dirtree{%
    .1 Glioblastoma Treatment Option Ontology.
    .2 Surgical Tumour Resection.
    .2 Radiotherapy.
    .3 Radiotherapy 60 Gy in 30 Fractions.
    .3 Radiotherapy 40 Gy in 15 Fractions.
    .2 Alkylating Agent Chemotherapy.
    .3 Temozolomide Chemotherapy.
    .3 Carmustine Chemotherapy.
    .4 Carmustine Polymer Wafer Implantation.
    .3 PCV Chemotherapy.
    .2 Lomustine (CCNU) Chemotherapy.
    .2 Alternating Electric Field Therapy.
    .2 Bevacizumab.
    }
    \caption{Subset of a glioblastoma treatment option ontology automatically constructed from the relevant clinical guidelines. Only the entities and the hierarchical relations are visualised without the corresponding text chunks and provenance relations. Note that the used LLM has incorrectly categorised Lomustine as a separate treatment rather than a variant of Alkylating Agent Chemotherapy, although this has no effect on the rest of the ArgEval pipeline. The nine leaves of the ontology are used in our main experiments.}
    \label{fig:gbm-ontology}
\end{figure}

The ontology is constructed in an iterative way, starting from a tuple of empty sets and being progressively expanded with newly appearing decision entities as well as links recording all entity mentions from the individual text chunks, as shown in Algorithm \ref{alg:ontology-construction}. The \texttt{mine\_ontology} function is implemented by an LLM provided with the current hierarchy of entities and the text of the current chunk in each iteration, and outputting data about new entities and entity mentions as part of a structured JSON. An example decision option ontology for the glioblastoma treatment recommendation task is shown in Figure \ref{fig:gbm-ontology}.

\subsubsection{General QBAF Construction} After extracting the decision-space ontology $\mathcal{O} = \langle \mathcal{E}, \mathcal{T}, \mathcal{H}, \mathcal{S} \rangle$ for a specific task, ArgEval uses this ontology to mine the general QBAFs $\mathbf{G} =\{\mathcal{G}_e\}_{e \in \mathcal{E}}$ for each identified option $e \in \mathcal{E}$. Each general QBAF is a tuple $\mathcal{G}_e = \langle \mathcal{A}_e, \mathcal{R}^-_e, \mathcal{R}^+_e, \tau_e, \chi_e \rangle$ where $\mathcal{A}_e, \mathcal{R}^-_e, \mathcal{R}^+_e, \tau_e$ denote arguments, a binary attack relation, a binary support relation and a base score function as for standard QBAFs (see Section \ref{sec:preliminaries}), while $\chi_e : \mathcal{A}_e \rightarrow \Phi$ is a condition function mapping arguments to formal conditions specifying the cases in which the given argument applies. $\Phi$ denotes the domain of conditions expressible in some suitable formal language.

Apart from the general QBAFs themselves, the procedure also produces a global parameter schema $\Pi$, containing the definitions of all parameters appearing in the extracted argument conditions. In our instantiation of the method, both the global parameter schema and the formalised conditions are expressed using JSON schemas,\footnote{\url{https://json-schema.org/}} which are relatively easy for LLMs to generate while establishing structure and value constraints that can be straightforwardly verified in an automated fashion. An example JSON schema formalising the condition for one of the arguments from Figure \ref{fig:argeval-inference} is given in Listing \ref{lst:json-condition}. Here, \texttt{eloquent\_structure\_involvement} and \texttt{kps} are parameters defined in the global parameter schema $\Pi$.

\begin{figure}[!tb]
\begin{lstlisting}[language=json,firstnumber=1,caption=JSON schema formalising the condition of the first attacking argument from Figure \ref{fig:argeval-inference}.,label=lst:json-condition,basicstyle=\fontfamily{fvm}\selectfont\small]
{
 "$schema": "https://json-schema.org/draft/2020-12/schema",
 "type": "object",
 "anyOf": [
  {
   "properties": {
    "eloquent_structure_involvement": {
     "type": "boolean",
     "const": true
    }
   }
  },
  {
   "properties": {
    "kps": {
     "type": "integer",
     "maximum": 49
    }
   }
  }
 ]
}
\end{lstlisting}
\end{figure}

The general QBAF construction procedure itself iterates over all the identified options $e \in \mathcal{E}$ in the ontology, proceeding in three main steps: bipolar argumentation framework mining, argument base score estimation and argument condition formalisation (see Algorithm \ref{alg:qbaf-construction}).

\noindent\textit{1) Bipolar Argumentation Framework Mining}. In the first step, the procedure employs an LLM to recursively mine arguments supporting and attacking the given decision option up to a specified depth $d$, forming a tree bipolar argumentation framework structure. This process is similar to the one used for ArgLLMs \citep{freedman-2025-argllms} with some important deviations: the LLM mining the arguments is also provided with all text chunks referring to the given decision, may be optionally supplied with an argument scheme $s_\text{arg}$ to guide the argument generation using domain-specific criteria (see an example in Table \ref{tab:arg-scheme}), can generate frameworks of an arbitrary breadth, with potentially different numbers of attackers and supporters, and produces natural-language conditions $\chi_e^N$ describing the circumstances under which each argument applies.

\noindent\textit{2) Argument Base Score Estimation}. Once the arguments have been generated, the second step of the construction process estimates their base scores. Depending on a boolean hyperparameter \texttt{score\_root}, this can either be done only for non-root arguments with the root receiving a midpoint score of $0.5$ or for all arguments in the framework including the root. The former option makes ArgEval predictions more sensitive to the influences of the root argument's attackers and supporters. Apart from the argument text itself, the scoring function is also provided with all text chunks associated with the option under consideration, and, for non-root arguments, the text of the parent argument, the relation of the evaluated argument to this parent and the natural-language condition associated with the argument. Similarly to ArgLLMs, our scoring function prompts an LLM to directly output the estimated base scores, as this simple strategy has been empirically found to outperform other, more complex approaches \citep{kevin}. This process results in base scores such as those shown in the example in Figure \ref{fig:argeval-inference}.

\begin{table}[tb]
\centering
\footnotesize
\begin{tabular}{p{2cm} p{5.5cm}}
\toprule
\textbf{Major premise} &
Generally, if a clinical intervention provides a net medical benefit and is comparatively suitable among its mutually exclusive alternatives for a given patient, then it is recommended. \\
\addlinespace
\textbf{Minor premise} &
The clinical intervention provides a positive benefit-risk balance for the considered patient. \\
\addlinespace
\textbf{Minor premise} &
The clinical intervention is superior or equivalent to its incompatible alternatives for the considered patient. \\
\midrule
\textbf{CQ1} &
Are the clinical features of the patient free of any contraindications against the considered intervention? \\
\addlinespace
\textbf{CQ2} &
Does the available evidence from clinical guidelines establish that the intervention provides a net benefit to the given patient? \\
\addlinespace
\textbf{CQ3} &
Is there no alternative intervention that offers a better benefit-risk profile for the given patient and is incompatible with the currently considered intervention? \\
\bottomrule
\end{tabular}
\caption{Premises and critical questions of the argument scheme $s_\text{arg}$ used for the glioblastoma treatment recommendation task. The scheme was developed specifically for this task in collaboration with a domain expert.}
\label{tab:arg-scheme}
\end{table}

\noindent\textit{3) Argument Condition Formalisation}. The final stage of the QBAF construction process converts the natural-language conditions $\chi_e^N$ associated with the arguments to the corresponding formal representations $\chi_e$. This procedure also iteratively updates the global parameter schema $\Pi$ with any new parameters referenced in the formalised conditions. Similarly to the other stages of the construction pipeline, this step leverages an LLM to generate each condition as well as the updated parameter schema.

After the construction process is finished, the obtained general QBAFs $\mathbf{G} =\{\mathcal{G}_e\}_{e \in \mathcal{E}}$ and the parameter schema $\Pi$ can be used for case-specific inference, as described in the following section.

\subsection{Case-Specific Inference}

\begin{algorithm}[tb]
\caption{Case-Specific Inference}
\label{alg:qbaf-inference}
\textbf{Input}: 
    A set of general QBAFs $\mathbf{G} = \{\mathcal{G}_e\}_{e \in \mathcal{E}}$, 
    global parameter schema $\Pi$, 
    natural-language case description $c$,
    argumentation semantics $\sigma$ \\
\textbf{Output}: 
    Sets of instantiated QBAFs $\mathbf{Q} = \{\mathcal{Q}_e\}_{e \in \mathcal{E}}$ and 
    decision recommendation scores $\mathbf{S} = \{s_e\}_{e \in \mathcal{E}}$
\begin{algorithmic}[1]
    \State Initialise $\mathbf{Q} \gets \emptyset$, $\mathbf{S} \gets \emptyset$
    
    \LineComment{1. Extract parameters from case description}
    \State $\mathcal{P} \gets \texttt{extract\_params}(c, \Pi)$
    
    \For{$\mathcal{G}_e \in \mathbf{G}$}
        \State $\langle \mathcal{A}_e, \mathcal{R}^-_e, \mathcal{R}^+_e, \tau_e, \chi_e \rangle \gets \mathcal{G}_e$
        
        \LineComment{2. Instantiate general QBAF into regular QBAF}
        \State Initialise $\mathcal{A}'_e \gets \mathcal{A}_e, \mathcal{R}'^-_e \gets \mathcal{R}^-_e, \mathcal{R}'^+_e \gets \mathcal{R}^+_e$
        
        \For{$a \in \mathcal{A}_e$}
            \LineComment{Remove arguments with unsatisfied conditions}
            \LineComment{and the associated relations}
            \If{$a \in \mathcal{A}'_e$ \textbf{and not} $\texttt{eval\_cond}(\chi_e(a), \mathcal{P})$}
                \State $\mathcal{A}_\text{rm} \gets \{a\} \cup \texttt{descendants}(a, \mathcal{R}'^-_e, \mathcal{R}'^+_e)$
                \State $\mathcal{A}'_e \gets \mathcal{A}'_e \setminus \mathcal{A}_\text{rm}$
                \State $\mathcal{R}'^-_e, \mathcal{R}'^+_e \gets \texttt{rm\_rels}(\mathcal{A}_\text{rm}, \mathcal{R}'^-_e, \mathcal{R}'^+_e)$
            \EndIf
        \EndFor
        
        \State $\tau'_e \gets \tau_e \text{ restricted to } \mathcal{A}'_e$
        \State $\mathcal{Q}_e \gets \langle \mathcal{A}'_e, \mathcal{R}'^-_e, \mathcal{R}'^+_e, \tau'_e \rangle$
        
        \LineComment{3. Calculate the recommendation score}
        \State $r_e \gets \texttt{get\_root}(\mathcal{Q}_e)$
        \State $s_e \gets \sigma_{\mathcal{Q}_e}(r_e)$
        
        \State $\mathbf{Q} \gets \mathbf{Q} \cup \{\mathcal{Q}_e\}$
        \State $\mathbf{S} \gets \mathbf{S} \cup \{s_e\}$
    \EndFor
    
    \State \Return $\mathbf{Q}, \mathbf{S}$
\end{algorithmic}
\end{algorithm}

The case-specific inference procedure of ArgEval first uses an LLM to extract a set of structured case parameters $\mathcal{P}$ from a natural-language description of the considered case $c$, guided by the definitions in the global parameter schema $\Pi$. For each general QBAF $\mathcal{G}_e \in \mathbf{G} $, these parameters can then be checked against the conditions associated with its arguments, instantiating these frameworks by removing any irrelevant arguments whose conditions are not satisfied, their descendants and any associated relations. This procedure yields regular, instantiated QBAFs $\mathbf{Q} = \{\mathcal{Q}_e\}_{e \in \mathcal{E}}$, which can be evaluated by a gradual semantics $\sigma$. The final strengths of the root arguments produced by the semantics can then be used as recommendation scores $\mathbf{S} = \{s_e\}_{e \in \mathcal{E}}$ quantifying the relative and absolute suitability of the associated decision options. An illustration of the inference process for a single decision option is shown in Figure \ref{fig:argeval-inference}, with the full procedure detailed in Algorithm \ref{alg:qbaf-inference}.

\subsection{ArgEval Properties}
\label{sec:argeval-properties}

ArgEval exhibits several useful properties summarised in Table \ref{tab:argeval-comparison}. Firstly, as we outlined above, the automatic decision-space ontology construction process used by ArgEval enables its application in \emph{open-ended domains} where the different decision options are not specified prior to applying the method. Additionally, since the predictions at inference time are derived by the deterministic argumentation semantics based on the instantiated QBAFs, these QBAFs can act as inherently \emph{faithful explanations} of the used reasoning process. This explainability and reliance on formal argumentative reasoning also facilitate \emph{local contestability}, enabling users to correct mistakes associated with inference for specific samples.

While the faithful explainability and local contestability properties have previously been considered and found to hold for other argumentative approaches \citep{freedman-2025-argllms,arg-rag}, ArgEval is unique in also supporting \emph{global contestability}. We consider a method to be globally contestable if it provides a faithfully interpretable mechanism $M$ such that contesting some aspect of $M$ for an instance $x$ can result in an updated mechanism $M'$, for which the resulting method performs better than the original on other instances $x' \neq x$ drawn from the same distribution, as measured by some pre-specified evaluation metric. That is, for a globally contestable method, user contestations on a specific input can persist and improve future predictions on different instances. ArgEval facilitates this through its reuse of general reasoning patterns captured in the decision-space ontology, general QBAFs and the general parameter schema, which can all be updated during contestation. These pre-compiled components also improve \emph{inference speed}, as case-specific inference only uses an LLM to extract the relevant case parameters, substantially reducing token usage.

\section{Experiments}
\label{sec:experiments}

Here, we describe the experiments applying ArgEval and additional baseline methods to the task of recommending suitable treatments for glioblastoma, an aggressive brain tumour with poor prognosis. Apart from reporting the initial performance on this task, we also conduct a case study illustrating the impact of contestation on the key evaluation metrics.

\subsection{Experimental Setup}

\paragraph{Task} In our experiments, we consider the task of providing patient-specific recommendations regarding suitable interventions for glioblastoma. The intention is to partially simulate the application of ArgEval and other baselines as a clinical decision support system advising clinicians on therapies that should be considered as part of the overall treatment plan. We believe this task poses a realistic and relevant evaluation setting, as the deployment of explainable AI systems would likely provide most substantial benefits in high-stakes domains such as healthcare.

\paragraph{Policy Documents} We use four established clinical guidelines on treating high-grade glioma or glioblastoma as our policy documents: the European Society for Medical Oncology (ESMO) clinical practice guidelines for high-grade glioma \citep{esmo-guidelines}, the National Comprehensive Cancer Network\textsuperscript{\textregistered} (NCCN\textsuperscript{\textregistered}) guidelines on central nervous system cancers\footnote{Disclaimer: NCCN makes no warranties of any kind whatsoever regarding their content, use or application and disclaims any responsibility for their application or use in any way.} \citep{nccn-guidelines}, the guideline on primary brain tumours and brain metastases in over 16s from the National Institute for Health and Care Excellence \citep{nice-guidelines}, and the Society for Neuro-Oncology (SNO) and European Society of Neuro-Oncology (EANO) consensus review on glioblastoma management in adults \citep{sno-eano-guidelines}. Due to the length of the selected guidelines, spanning more than $200$ pages in one case, we manually extracted pages relevant to glioblastoma treatment.

To parse and preprocess the unstructured guideline PDFs, we used the Docling library \citep{docling} along with its heron document layout analysis model \citep{livathinos-2025-layout} and a hybrid chunker splitting the documents into shorter chunks. These chunks were postprocessed by gpt-oss-20b \citep{gpt-oss-20b}, merging semantically related chunks that were separated by page breaks, figures or other PDF elements and filtering out chunks unrelated to glioblastoma treatment. This process resulted in a total of 67 chunks.

Given the refined chunks, we used gpt-oss-20b to construct a decision-space ontology using the procedure described in Section \ref{sec:argeval}, identifying a total of 179 treatment entities. To constrain our experiments to the most common atomic treatment options, we only retained leaf entities associated with mentions in at least three out of the four considered guideline documents that were not combination treatments (i.e., they had a single root ancestor in the ontology hierarchy). This yielded an ontology with 9 different treatment options captured in Figure \ref{fig:gbm-ontology}. A review by an expert oncologist determined that the ontology captures the most commonly considered non-combination therapies for glioblastoma, along with some less relevant options, providing a useful basis for the subsequent experiments.

\paragraph{Case Descriptions} To source the patient case descriptions, we again liaised with an expert clinical oncologist to obtain a set of four key parameters that substantially affect treatment decisions for glioblastoma patients along with suitable value ranges: age (ranging over the values 50, 60, 75 and 85), MGMT methylation status (ranging over the values ``methylated'', ``unknown'' and ``unmethylated''), Karnofsky Performance Status (KPS, ranging over the values 10, 30, 50, 70 and 90) and tumour location (ranging over the values ``non-dominant frontal lobe'', ``thalamus'' and ``brainstem''). In addition to these properties, we also add a spurious parameter indicating the patient sex (``male'' or ``female''), which is typically seen as having no impact on the choice of appropriate therapy for glioblastoma. Considering all possible combinations of parameter values provides us with $4 \times 3 \times 5 \times 3 \times 2 = 360$ unique patient parameter sets, which we then convert into brief patient vignettes using an LLM (see Figure \ref{fig:argeval-inference} for an example). We performed manual checks to ensure that the vignettes accurately represent the input patient parameters with no omissions or extraneous information.

\paragraph{Labels} The ground-truth treatment recommendation labels were determined according to an algorithm with rules informed and reviewed by the clinical oncologist, aiming to capture the general clinical practice when deciding on treatments recommended to different patient subgroups. The labels are assigned to each patient vignette/treatment combination, resulting in $360 \times 9 = 3240$ labels in total. Each label indicates whether a specific treatment is ``recommended'', ``maybe recommended'' or ``not recommended'' for a patient with the given characteristics, with the ``maybe recommended'' label indicating potential suitability with relatively weak or conflicting evidence.

\begin{table*}[t]
\centering
\caption{Performance on glioblastoma (GBM) treatment prediction across models and inference configurations. LMR denotes the label match rate, indicating the share of the absolute treatment recommendation scores matching the labels, while NDCG quantifies the level of agreement between the ground-truth treatment ranking and the predicted ranking. The I/O Tokens column reports prompt/completion token usage in millions (M) without the initial decision-space ontology construction, which is shared for all methods.}
\label{tab:main-results}
\begin{tabular}{llcccccc}
\toprule
LLM Backbone & Method & $d$ & Est. Root & Arg. Scheme & LMR$^\uparrow$ & NDCG$^\uparrow$ & I/O Tokens (M)$^\downarrow$ \\
\midrule

\multicolumn{8}{l}{\textbf{gpt-oss-20b (medium)}} \\
\midrule

gpt-oss-20b & Base LLM   & -- & --    & --    & 0.8775 & 0.9578 & 16.95 / 6.59 \\

\addlinespace
gpt-oss-20b & ArgLLMs-O  & 1  & \xmark & \xmark & 0.8614 & 0.9739 & 80.51 / 20.12 \\
gpt-oss-20b & ArgLLMs-O  & 1  & \xmark & \cmark  & 0.8676 & 0.9629 & 80.41 / 19.89 \\
gpt-oss-20b & ArgLLMs-O  & 1  & \cmark  & \xmark & 0.8429 & 0.9718 & 95.91 / 22.69 \\
gpt-oss-20b & ArgLLMs-O  & 1  & \cmark  & \cmark  & 0.8444 & 0.9707 & 95.98 / 22.72 \\

\addlinespace
gpt-oss-20b & ArgEval & 1  & \xmark & \xmark & 0.6910 & 0.9330 & 1.18 / 0.79 \\
gpt-oss-20b & ArgEval & 1  & \xmark & \cmark  & 0.8009 & 0.9654 & 1.20 / 0.62 \\
gpt-oss-20b & ArgEval & 1  & \cmark  & \xmark & 0.4432 & 0.9293 & 1.18 / 0.66 \\
gpt-oss-20b & ArgEval & 1  & \cmark  & \cmark  & 0.6923 & 0.8983 & 1.15 / 0.68 \\

\addlinespace
gpt-oss-20b & ArgEval & 2  & \xmark & \xmark & 0.8485 & 0.9422 & 3.96 / 1.93 \\
gpt-oss-20b & ArgEval & 2  & \xmark & \cmark  & 0.7006 & 0.8514 & 4.09 / 2.02 \\
gpt-oss-20b & ArgEval & 2  & \cmark  & \xmark & 0.5244 & 0.9480 & 4.09 / 1.92 \\
gpt-oss-20b & ArgEval & 2  & \cmark  & \cmark  & 0.6914 & 0.9241 & 4.51 / 2.29 \\

\midrule

\multicolumn{8}{l}{\textbf{Qwen3-30B-A3B-FP8}} \\
\midrule

Qwen3-30B & Base LLM    & -- & --    & --    & 0.8630 & 0.9527 & 17.83 / 5.44 \\

\addlinespace
Qwen3-30B & ArgLLMs-O  & 1  & \xmark & \xmark & 0.8373 & 0.9752 & 78.06 / 22.11 \\
Qwen3-30B & ArgLLMs-O  & 1  & \xmark & \cmark  & 0.8617 & 0.9735 & 78.95 / 21.73 \\
Qwen3-30B & ArgLLMs-O  & 1  & \cmark  & \xmark & 0.8216 & 0.9815 & 93.93 / 25.80 \\
Qwen3-30B & ArgLLMs-O  & 1  & \cmark  & \cmark  & 0.8417 & 0.9804 & 96.06 / 25.75 \\

\addlinespace
Qwen3-30B & ArgEval & 1  & \xmark & \xmark & 0.8812 & 0.9750 & 0.90 / 0.67 \\
Qwen3-30B & ArgEval & 1  & \xmark & \cmark  & 0.8623 & 0.9389 & 0.89 / 0.67 \\
Qwen3-30B & ArgEval & 1  & \cmark  & \xmark & 0.7417 & 0.9138 & 0.88 / 0.78 \\
Qwen3-30B & ArgEval & 1  & \cmark  & \cmark  & 0.7497 & 0.9118 & 1.01 / 0.55 \\

\addlinespace
Qwen3-30B & ArgEval & 2  & \xmark & \xmark & 0.7886 & 0.9782 & 2.31 / 1.38 \\
Qwen3-30B & ArgEval & 2  & \xmark & \cmark  & 0.8818 & 0.9771 & 2.65 / 1.38 \\
Qwen3-30B & ArgEval & 2  & \cmark  & \xmark & 0.6284 & 0.8957 & 2.36 / 1.36 \\
Qwen3-30B & ArgEval & 2  & \cmark  & \cmark  & 0.7694 & 0.9123 & 3.01 / 1.73 \\

\bottomrule
\end{tabular}
\end{table*}

\paragraph{LLM Backbones} We consider two reasoning LLMs from two different providers, gpt-oss-20b \citep{gpt-oss-20b} and Qwen3-30B-A3B-FP8 \citep{qwen-3}. Both models rely on the Mixture-of-Experts (MoE) architecture for more efficient inference, with 3.6B and 3.3B active parameters, respectively. We selected these models due to their high scores on the Artificial Analysis Intelligence Index\footnote{\url{https://artificialanalysis.ai/\#intelligence}} among major open-weight models that fit within our resource constraints. In our experiments, both models were run in reasoning mode, setting the reasoning effort level to ``medium'' for gpt-oss-20b. Inference was performed using the vLLM engine \citep{kwon-2023-vllm} with the sampling temperature set to $1.0$ and other generation parameters taking their default values. We opted to use this temperature value since it is the recommended sampling temperature for gpt-oss-20b and since high temperatures are often needed to avoid degenerate behaviours in current reasoning LLMs, particularly looping \citep{pipis-2025-reasoning-models-loop}. While the main experiments use both considered models, the initial document processing and ontology construction were only performed using gpt-oss-20b, ensuring that all the experimental results are based on a consistent ontology.

\paragraph{Baselines} We compare our method to two baselines: base LLMs and ArgLLMs-O. Base LLMs are directly instructed to provide a numerical score ranging from 0 to 1 indicating the degree to which a treatment is recommended for a given patient, with this process being repeated for each of the 9 treatment options. The LLM is provided with all the text chunks associated with the currently evaluated treatment option as well as the patient case description, and can use the reasoning chain for intermediate work before giving the final answer. The ArgLLMs-O baseline is an adapted version of the original ArgLLMs \citep{freedman-2025-argllms} using our treatment option ontology and generating a separate QBAF for each possible treatment option. During both the argument generation and base score estimation stages, the LLM performing these operations is provided with all the relevant guideline chunks as well as the patient case description. Thus, both baselines can use the patient information during their entire reasoning process, while ArgEval only gains access to it during the case-specific inference stage.

\begin{figure*}[tb]
    \centering
    \includegraphics[width=0.95\textwidth]{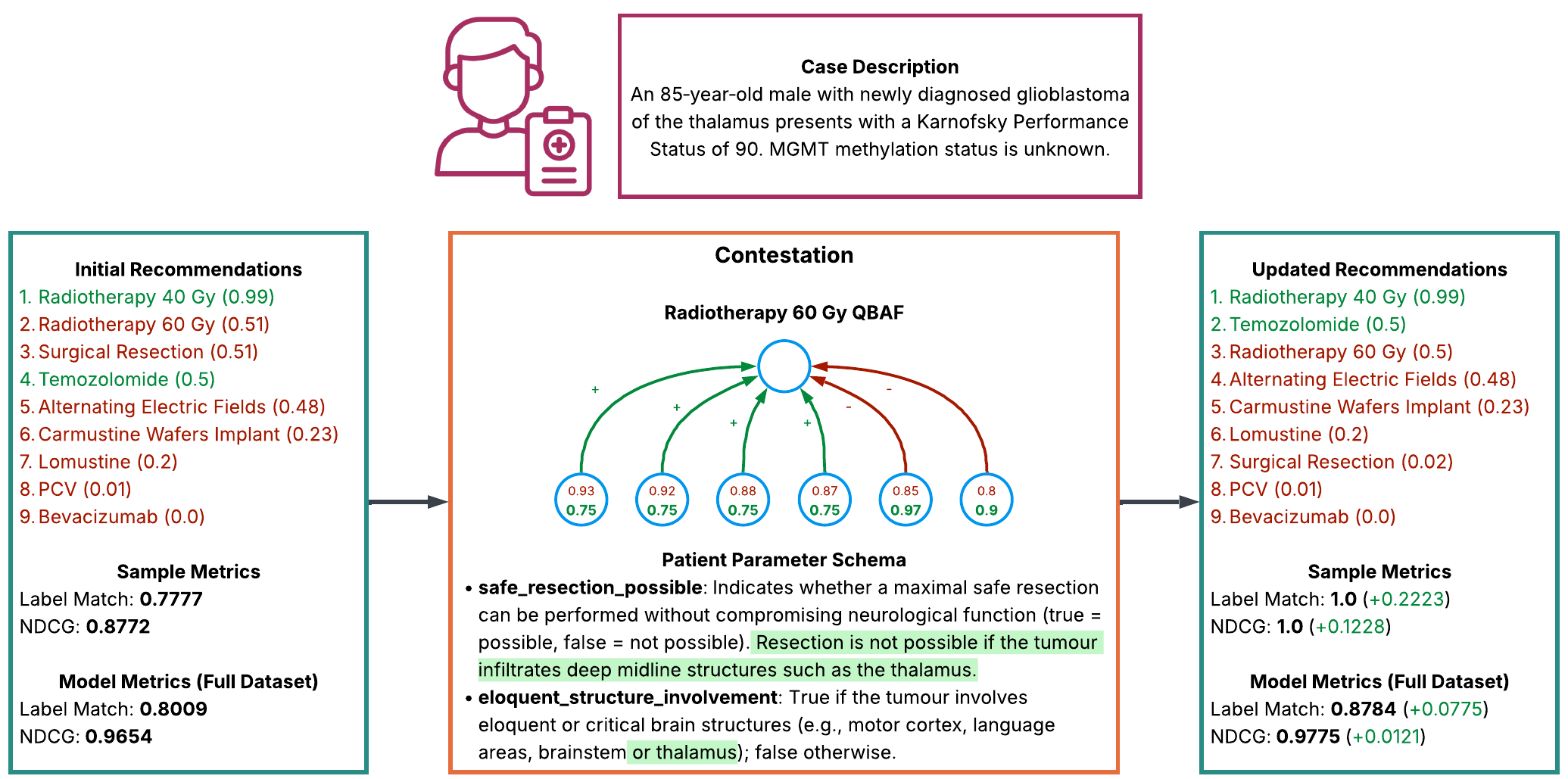}
    \caption{Illustration of the ArgEval contestability experiment\protect\footnotemark. To correct the initially suboptimal recommendations, we make small adjustments to the base scores in the general argumentation framework for the radiotherapy 60 Gy treatment option (with initial scores shown in smaller red font and updated scores in larger green font) and clarify the descriptions of two parameters associated with surgical resection in the parameter schema. These modifications are sufficient to achieve a perfect score on this instance while also substantially improving the overall performance. Treatments recommended by the ground-truth labels are shown in green, with those not recommended in red.}
    \label{fig:argeval-contestability}
\end{figure*}

\paragraph{Method Variants} We consider eight different variants of ArgEval arising from the choices of three hyperparameters: the general QBAF depth (either 1 or 2), root base score estimation (either enabled or disabled) and argument scheme inclusion (either enabled or disabled). The maximum QBAF depth is limited to two, as considering larger depths would likely provide diminishing returns while considerably increasing computational costs and cognitive load on the users. See Section \ref{sec:argeval}, Algorithm \ref{alg:qbaf-construction} and Table \ref{tab:arg-scheme} for more information on these parameters. For ArgLLMs-O, we similarly consider variations with or without root base score estimation and argument scheme inclusion. Due to the very high computational requirements of ArgLLMs-O, we only consider depth 1 versions for this baseline. All argumentative models use the DF-QuAD gradual semantics \citep{RagoTAB16}, which is also the default semantics for ArgLLMs.

\paragraph{Metrics} We use three key metrics to compare the performance of the different methods. The label match rate (LMR) indicates the share of the absolute treatment recommendation scores matching their labels, with the ``recommended'' label considered to be matching for scores between $0.5$ and $1.0$, the ``maybe recommended'' label matching scores between $0.25$ and $0.75$ and the ``not recommended'' label matching scores between $0.0$ and $0.5$. Note that the used score intervals intentionally overlap in order to reflect inherent uncertainty associated with decision-making in medicine. This design choice is also intended to reduce the sensitivity of the metrics to the precise score scales used by different methods. Since the recommendation scores are intended to be helpful in ranking potential treatments from the most to least suitable, we also report the normalised discounted cumulative gain (NDCG) metric \citep{jarvelin-2002-ndcg}. NDCG indicates the level of agreement between the ground-truth ranking (starting from the ``recommended'' treatments and ending with the ``not recommended'' ones) and the ranking defined by the predicted treatment recommendation scores. Finally, we report the total prompt and completion tokens used during the LLM inference for each method. These measures include the general QBAF construction for ArgEval but exclude the initial ontology mining, which is shared for all methods.

\subsection{Performance Results}
The results of our experiments are summarised in Table \ref{tab:main-results}. There are several trends that can be observed.

Generally, the best-performing versions of ArgEval achieve competitive results compared to the baselines. For example, the variant using the Qwen3-30B model with depth 2, no root estimation and argument scheme inclusion achieves the best overall LMR of $0.8818$ with a comparatively high NDCG of $0.9771$, which is only surpassed by 3/25 other method variants (one of which is another instance of ArgEval). However, ArgEval seems to be more sensitive to the chosen hyperparameters with substantial variations between the different versions. This is likely due to the fact that any deficiencies during general QBAF construction affect all subsequent inferences. ArgEval variants using root base score estimation perform particularly poorly, probably because treatment options assigned high base scores may still be inappropriate for specific patients.

Nevertheless, the high performance of certain ArgEval variants is notable given their low computational requirements. In particular, thanks to its global reasoning, ArgEval requires substantially fewer inference tokens compared to the other baselines, with even the most expensive depth 2 variant requiring $\sim$$2.4\times$ and  $\sim$$8.7\times$ fewer completion tokens compared to the cheapest base LLM version and the cheapest ArgLLMs-O version, respectively.

\footnotetext{Icon from Noun Project (\href{https://thenounproject.com/icon/patient-6651912/}{Patient} by Alzam, \href{https://creativecommons.org/licenses/by/3.0/deed.en}{CC BY 3.0}).}

Overall, we argue that ArgEval provides a good trade-off between recommendation performance and computational costs, with a substantial qualitative benefit associated with global contestability. In the next section, we demonstrate that this benefit can translate into direct performance gains.

\subsection{Contestability Case Study}

To demonstrate the explainability and global contestability capabilities of ArgEval, we conduct an additional experiment showing the performance effects of contesting the recommendations for a single output. We intentionally choose an ArgEval variant with moderate performance, specifically, the version using gpt-oss-20b with depth 1, no root score estimation and argument scheme inclusion.

As illustrated in Figure \ref{fig:argeval-contestability}, the model correctly recommends Radiotherapy 40 Gy, but incorrectly ranks the unsuitable options Radiotherapy 60 Gy and Surgical Resection above Temozolomide, another possible therapy. Inspecting the instantiated QBAFs for Radiotherapy 60 Gy, we observe that there are four supporting and two attacking arguments. While the attacking arguments correctly identify Radiotherapy 40 Gy as typically being the more appropriate regimen for elderly patients, they are outweighed by the supporters arguing that Radiotherapy 60 Gy is generally suitable for patients with good performance status. Aiming to perform a minimal contestation correcting the recommendations, we slightly increase the base scores of the attackers and decrease the base scores of the supporters in the general QBAF, reducing the Radiotherapy 60 Gy score to slightly below 0.5 (not visible in Figure \ref{fig:argeval-contestability} due to rounding).

For Surgical Resection, our inspection of the instantiated QBAFs reveals that ArgEval has correctly captured conditions associated with surgery contraindications, but is unable to infer that these contraindications apply to patients with glioblastoma of the thalamus during parameter extraction. To address this issue, we refine the corresponding parameter descriptions in the schema as illustrated. This results in correct extraction of the associated parameters and a substantial reduction in the Surgical Resection score, with the inference process after these changes illustrated in Figure \ref{fig:argeval-inference}.

Notably, performing the two contestations based on a single sample results in a substantial performance improvement on the full dataset, reaching an LMR of $0.8784$ and NDCG of $0.9775$, which surpasses the performance of all other methods using gpt-oss-20b.

\section{Conclusion}

In this paper, we introduced ArgEval, a novel framework automatically mapping task decision spaces, mining the corresponding option ontologies and constructing general QBAFs capturing the benefits and drawbacks of each decision option depending on the characteristics of a specific case. Uniquely, this shift to general argumentative reasoning enables global contestability while also providing substantial computational cost savings. We evaluated the performance of ArgEval on the task of treatment decision recommendation for glioblastoma, finding it to be highly competitive, with contestability providing opportunities for further performance improvements. Future work could further explore the utility of ArgEval's explainability and contestability to end users through user studies or consider its application in domains beyond healthcare.

\section*{Acknowledgments}
Dejl and Toni were partially funded by the European Research Council (ERC) under the European Union’s Horizon 2020 research and innovation programme (grant agreement No. 101020934, ADIX). Toni was also funded by EPSRC (grant UKRI3928, NeSyDebates). Williams was partially funded by NIHR Imperial BRC (grant no. RDB01 79560) and by the Brain Tumour Research Centre of Excellence at Imperial.

\bibliography{references}

\newpage
\appendix

\end{document}